\documentclass[letterpaper, 10 pt, conference]{ieeeconf}  

\IEEEoverridecommandlockouts                              

\usepackage{amsmath,amsfonts}
\usepackage{algpseudocode} 
\usepackage{algorithm}
\usepackage{array}
\usepackage[caption=false,font=normalsize,labelfont=sf,textfont=sf]{subfig}
\usepackage{textcomp}
\usepackage{stfloats}
\usepackage{url}
\usepackage{verbatim}
\usepackage{graphicx}
\usepackage{cite}
\usepackage{xcolor} 
\usepackage{subcaption} 

\title{\LARGE \bf
Towards Soft Robotic Exogloves for Musculoskeletal Manipulation to Reduce Pain and Spasticity
}

\author{Antonia Salluce$^{1,2}$, Maeryn Erdheim$^{1,2}$, Gailen Davis$^{1}$, Lauren H. Sullivan$^{1}$, Max-William Kanz$^{1}$,\\and Jacqueline Libby$^{3}$,~\IEEEmembership{Member,~IEEE}
\thanks{This work was supported by the US National Science Foundation under Grant 2502197. (\emph{Corresponding author: Jacqueline Libby})}%
\thanks{$^{1}$Antonia Salluce, Maeryn Erdheim, Gailen Davis, Lauren H. Sullivan, and Max-William Kanz are with the Department of Mechanical Engineering, Stevens Institute of Technology, 1 Castle Point Terrace, Hoboken, NJ, 07030, USA
        {\tt\small asalluce@stevens.edu, merdheim@stevens.edu, gdavis3@stevens.edu, lsulliv1@stevens.edu, mkanz@stevens.edu}}%
\thanks{$^{2}$Antonia Salluce and Maeryn Erdheim contributed equally to this article.}%
\thanks{$^{3}$Jacqueline Libby is with Faculty of Mechanical Engineering, Stevens Institute of Technology, 1 Castle Point Terrace, Hoboken, NJ, 07030, USA
        {\tt\small jlibby@stevens.edu}}%
}

\begin{document}

\maketitle
\thispagestyle{empty}
\pagestyle{empty}

\begin{abstract}
Hand spasticity and resulting pain affect 12 million people worldwide, including stroke survivors, arthritis patients, and those with other muscle and nerve deficiencies.
Soft robotic exogloves are being introduced to help patients enhance mobility or manage pain; however, there are no current solutions that address both pain and mobility.
We present preliminary development of a soft robotic exoglove that both aids in mobility and administers massage-like compression to relax spastic muscles. 
The glove consists of soft pneumatic actuators that are personalized to an individual's hand topology and kinematics, allowing for optimal conformability and targeted mobility.
Novel soft actuators were designed, analyzed, fabricated, assembled into an exoglove, and experimentally tested.
Actuators were 3D modeled and analyzed with finite element modeling under pressures of 100 and 200 kPa.
Geometries were optimized to minimize stress before fabrication and testing.
A dorsal finger actuator was successfully customized to a participant's hand topology, providing full conformal contact and maximal force distribution.
A ventral finger actuator was successfully fabricated that can be drastically compressed in size to fit into the tight space of a hyperflexed spastic finger.
A palmar actuator was successfully printed with stereolithography, showing potential for 3D-printed soft actuators with more complex geometries.
The glove was assembled and successfully worn by a pilot user to validate initial findings in comfort and effectiveness.
\end{abstract}

\section{INTRODUCTION}\label{sec.introduction}

Spasticity is a large contributor to motor impairment in individuals with neurological disorders, greatly limiting daily function and quality of life. A recent meta-analysis was conducted that revealed 25.3\% of stroke patients develop spasticity~\cite{zeng2021prevalence}. Spasticity also occurs frequently in other neurological disorders, with prevalence ranging from 20-83\% in patients with spinal cord injury, 50-66\% in patients with multiple sclerosis, and 75-90\% in children with cerebral palsy~\cite{therkildsen2025evaluation}. The advancement of rehabilitation devices has sought to augment or complement traditional physical therapy. Specifically, soft, wearable robotics are promising in improving motor function and enabling repetitive motion training in spastic populations. This potential is evidenced by the development of soft robotic gloves for both assistive and at-home hand rehabilitation~\cite{polygerinos2015soft}. 

Within this class of soft, wearable systems, cable-driven robotic exogloves have become a widely adopted solution in assisting with the opening and closing of the hand for stroke rehabilitation. Cable-driven hand rehabilitation devices typically employ tendon or Bowden cable transmission routed along the fingers to assist with flexion and extension~\cite{xiloyannis2016modelling, mohammadi2018flexo, kim2018cable, liu2023finger}.
However, friction in the cables can introduce resistance, potentially limiting the effectiveness of force transmission~\cite{alicea2021soft}. Because actuation forces in cable-driven rehabilitation devices are applied through discrete attachment locations, achieving uniform force distribution is also challenging~\cite{galiana2012wearable} and can reduce comfort~\cite{liu2023finger}. Furthermore, while underactuated cable-driven designs simplify the system, their reliance on limited degrees of freedom may restrict natural hand motion, which is problematic in rehabilitation contexts~\cite{xiloyannis2016modelling}. 

Soft elastomeric pneumatic actuators have become a prominent actuation approach in soft wearable robotic gloves due to their ability to conform to complex anatomy and distribute force across the surface of the hand. The work by Heung et al. presents a soft robotic glove that enables finger flexion and extension for assistive purposes~\cite{heung2019robotic}. Walsh’s group demonstrated pneumatic actuation using Pneu-Net actuators to produce controlled bending motions for finger rehabilitation~\cite{polygerinos2013towards}. In addition, Ridremont et al. developed a pneumatically actuated device capable of assisting both finger and wrist motion, assisting activities of daily living (ADL) tasks~\cite{ridremont2024pneumatically}. Despite these advancements, these pneumatic glove designs focus primarily on improving hand mobility, strength, and activities of daily living, with few addressing pain reduction or joint stiffness mitigation.

Soft robotic systems employing reinforced fabric pneumatic actuators have been widely explored for portable, lightweight compression and massage therapy, particularly for lower-limb applications. These devices leverage relatively simple fabrication processes and enable customized fits through fabric-based actuator geometries~\cite{rosalia2021soft}. Rather than using molded silicone actuators, fabric pockets or strap-based pneumatic chambers can be sequentially inflated along the limb to generate controlled compression and personalized massage patterns~\cite{zhu2023peristaltic}. Such systems are especially beneficial for individuals who have difficulty donning traditional compression garments, as the actuators initially accommodate larger volumes before contracting to conform closely to the limb. Fabric-based pneumatic actuators have demonstrated the ability to deliver clinically relevant levels of compression, supporting their use in therapeutic and rehabilitative contexts.

Fabric-based pneumatic and hydraulic soft robotic actuators have also been extended to upper-body applications, including the hand, wrist, and arm. Cyclic pneumatic compression applied to the fingers has been shown to effectively reduce joint pain and stiffness, particularly for individuals with rheumatoid arthritis~\cite{chua2019design}. To improve wearability and comfort, actuators can be embedded directly within textile structures, enabling programmable, human-like massage through sequential actuation~\cite{kim2023knitdema}. Beyond pneumatic systems, hydraulic soft actuators have been used to generate peristaltic compression along the upper arm, allowing for personalized and distributed therapy~\cite{zhu2023peristaltic}. Other soft robotic approaches, such as thin McKibben muscle-based gloves, provide hand rehabilitation by initially taking the shape of a high-volume sleeve configuration and subsequently contracting to stabilize finger joints, promote circulation, and alleviate pain~\cite{koizumi2020soft}. While robotic-assisted hand rehabilitation combined with conventional therapy has demonstrated reductions in paralysis and associated symptoms~\cite{borboni2017robot}, many existing solutions primarily target motor recovery and functional assistance or pain management alone, with limited emphasis on sustained pain management and rehabilitation in one. Furthermore, non-modular devices resembling traditional gloves often present challenges related to taking devices on and off, particularly for users with reduced hand mobility, which remains a key limitation for widespread adoption in therapeutic applications of the hand.

Building upon our prior work~\cite{Libby2023-ISMR,Libby2023-arXiv,Massoud2024-IEEEAccess,Massoud2025-RAL}, 
our contribution focuses on the design and fabrication of key components for a custom, modular soft robotic glove that addresses the therapeutic needs of patients with hand spasticity, stiffness, and pain. Our modular glove includes silicone-casted pneumatic actuators for the dorsal and ventral surfaces of the finger, as well as the palm of the hand, that can apply compressive, distributed forces when inflated with a pneumatic control system. By adjusting the pressures of each actuator, the hand can be gradually repositioned, compressed, and massaged to each patient’s comfort level and therapeutic needs. 
The distributed forces provided by the actuators in the glove could also be used for resistance in strength training and assistance for neuroplasticity training, in parallel with spasticity reduction. 
The ventral and palmar actuators are designed with compressible bellows that can be fit into the tight spaces of a contracted hand, allowing the glove to be correctly positioned and more easily donned and doffed.
Furthermore, topological customization of the actuators is presented, which allows for highly personalized gloves that can potentially provide optimal pressure distribution and targeted musculoskeletal manipulation. 
This personalization is a step towards a future class of soft wearable devices that 
will enable on-demand, long-lasting pain relief as well as complementary rehabilitation sessions to in-person sessions at the clinic, hopefully resulting in a significant increase in recovery from spasticity. 
\section{METHODS}\label{sec.methods}
\begin{figure}[h]
\centering
\includegraphics[width=1\linewidth]{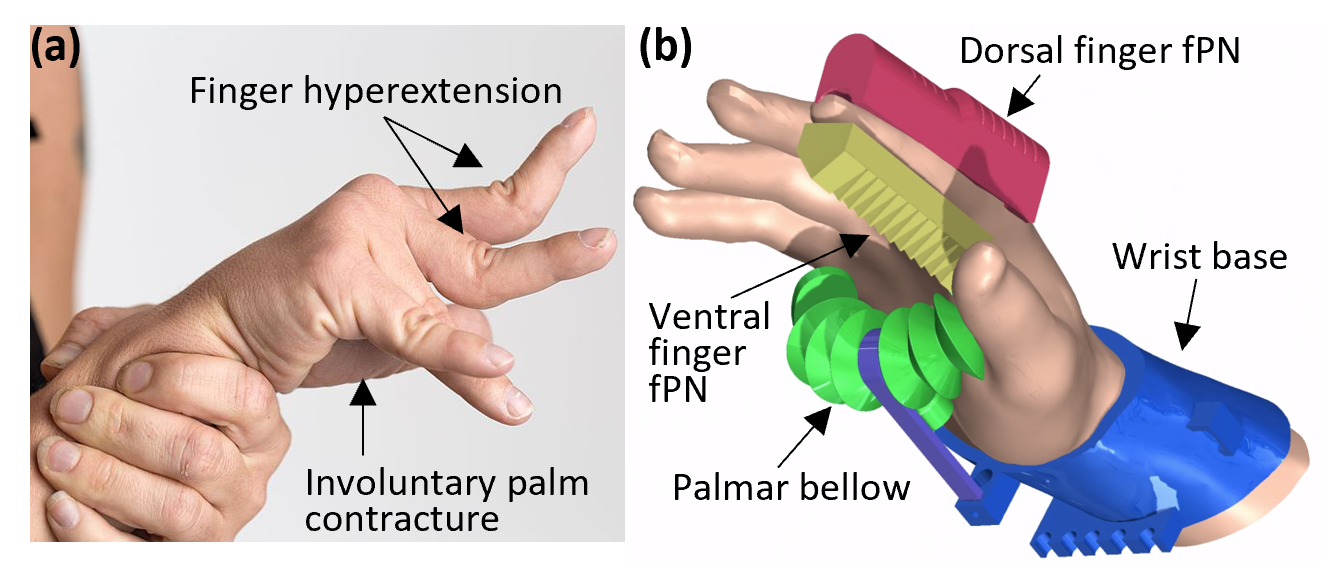}
\caption{(a) Stroke survivor with spastic hand~\cite{website-physiopediaSpasticity}.
(b) CAD of the presented soft robotic exoglove.}
\label{fig.spasticAndProto}
\end{figure}

We present the design and fabrication of a soft robotic exoglove for massage, compression, and repositioning of a spastic hand to relax the muscles and encourage a neutral hand posture.
Fig.~\ref{fig.spasticAndProto}a shows a stroke survivor with a spastic hand suffering from involuntary palm contracture and finger hyperextesnion~\cite{website-physiopediaSpasticity}. 
Fig.~\ref{fig.spasticAndProto}b is the CAD assembly of the presented soft robotic exoglove,
consisting of three soft actuators fabricated from silicone and pneumatically controlled.
The actuation occurs through air tubes connected to a robotically-operated desktop pneumatic control station presented in our previous work~\cite{Massoud2024-IEEEAccess, Massoud2025-RAL}.
The dorsal (pink) and ventral (yellow) actuators can be 
pressurized individually to encourage finger repositioning into flexion or extension, respectively,
or can be
pressurized simultaneously to apply compression to the finger. 
The palmar actuator (green) is pressurized to apply distributed forces to the thenar and hypothenar eminences to massage a contracted palm and encourage palm repositioning.

Both the external and internal geometries of all three soft actuators are customized for each individual's hand topology.
The dorsal and ventral finger actuators are extensions of a traditional fast pneunet (fPN) soft robotic design~\cite{Mosadegh2014-AdvFunctMater}, and the palmar actuator is an extension of a traditional bellow design~\cite{hu2020novel}.
The geometries of the ventral finger fPN and the palmar bellow allow for elongation and compression of the actuator. 
In particular, compression of an actuator allows it to decrease in size so that it can be fit into the tight spaces of a contracted palm or flexed finger.

Section~\ref{subsec.customizedDorsalFpn} presents the design of the dorsal finger fPN and the scanning process used to achieve a personalized topology for each individual.
\ref{subsec.ventralfPN} presents the design of the ventral finger fPN.
\ref{subsec.fPNFabrication} presents the casting fabrication of the dorsal and ventral finger fPNs.
\ref{subsec.palmarBellow} presents the design of the palmar bellow. (The 3D printing of the bellow will be discussed in Section~\ref{sec.experimentsAndResults}.)
\ref{subsec.wristBase} presents the wrist base.

\subsection{Customized Dorsal fPN}\label{subsec.customizedDorsalFpn}

\begin{figure}[h]
\centering
\includegraphics[width=1\linewidth]{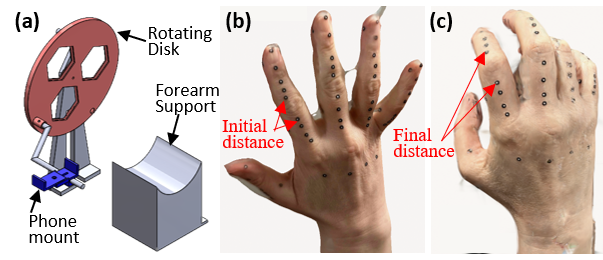}
\caption{Photogrammetry cans with (a) relaxed hand with fingers in full flexion, and (b) spastic hand with bending at joints.}
\label{fig.crankAndPolycam}
\end{figure}

The geometry of the hand was captured using photogrammetry.
An iPhone was mounted on a custom-built hand-crank mechanism as shown in Fig.~\ref{fig.crankAndPolycam}a.
The individual's forearm rested on the forearm support to keep the hand stationary, while a 25-second video was recorded during a 360$^\circ$ rotation.
The video was uploaded to Polycam to generate a 3D surface mesh of the hand.
Two scans were collected with fingers extended (Fig.~\ref{fig.crankAndPolycam}b) and flexed (Fig.~\ref{fig.crankAndPolycam}c).


The mesh is post-processed in Meshmixer (Autodesk, Inc) where it is denoised, cropped, scaled, and turned into a solid body.
The known diameter (3mm) of the markers stuck to the hand is used to scale the model, taking an average across multiple markers.

\begin{figure}[h]
\vspace{2pt}
\centering
\includegraphics[width=1\linewidth]{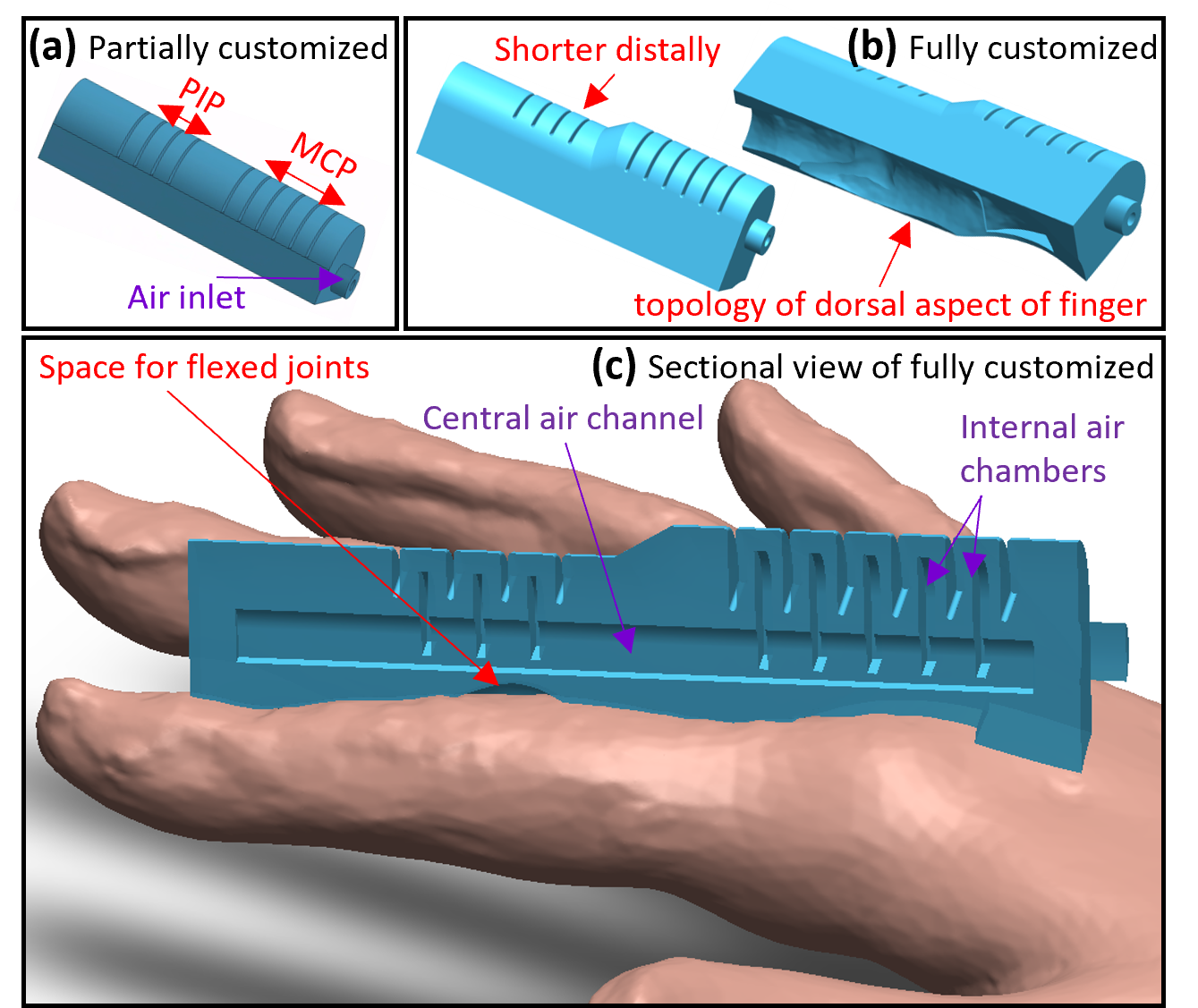}
\caption{Customized dorsal fPN actuator. \emph{(purple)} Pneumatic features. \emph{(red)} customization features. (a) Traditional fPN, partially customized with air chambers centered at the MCP and PIP joint locations. (b) Fully customized with conformal base topology and shorter distal portion. (c) Sectional view of fully customized fPN on the participant’s hand, depicting the internal actuator geometry, as well as the extra space at the joints for unrestricted finger flexion.}
\label{fig.dorsalFPN}
\end{figure}

The dorsal fPN (Fig.~\ref{fig.spasticAndProto}b pink) fits on the dorsal aspect of the finger.
When inflated with air, it applies pressure to the finger to encourage flexion, as well as compression, which promotes muscle relaxation and blood flow.
The novel dorsal fPN design, presented in Fig.~\ref{fig.dorsalFPN}, is customized to an individual's scan in several ways.
First, there are alternating bending and straight regions,
with bending regions centered around the metacarpophalangeal (MCP) and proximal interphalangeal (PIP) joints, as depicted in the more traditional but partially customized fPN in Fig.~\ref{fig.dorsalFPN}a.
The bending is caused by internal air chambers, whereas the straight regions are solid silicone with only an air channel.
The length of each bending region is determined from the initial distance labeled in Fig.~\ref{fig.crankAndPolycam}b, with the hand extended. 
The second customization, shown in Fig.~\ref{fig.dorsalFPN}b \emph{left}, is a decrease in the height of the fPN toward the distal tip of the finger.
Third, the fPN bottom surface (Fig.~\ref{fig.dorsalFPN}b \emph{right}), which contacts the finger, is molded to match the dorsal finger topology, enabling conformal contact.
By maximizing contact, the pressure from the actuator can be distributed more evenly.
The dorsal fPN model is modified by subtracting the solid model of the hand from the model.
Fourth, additional space was added at the joint contact surfaces to accommodate the increased dorsal finger surface area during flexion.
The differences in initial and final distances around the PIP joint when the finger flexes can be seen in Fig.~\ref{fig.crankAndPolycam}(b,c). 
Hence, the flexed hand scan was processed and subtracted from the fPN model to create this additional space for the PIP joint. In the future, the same will be done for the MCP joint.
This space is denoted with a red arrow in the profile sectional view in Fig.~\ref{fig.dorsalFPN}c.

\subsection{Ventral fPN}\label{subsec.ventralfPN}

\begin{figure}[h]
\vspace{6pt}
\centering
\includegraphics[width=.9\linewidth]{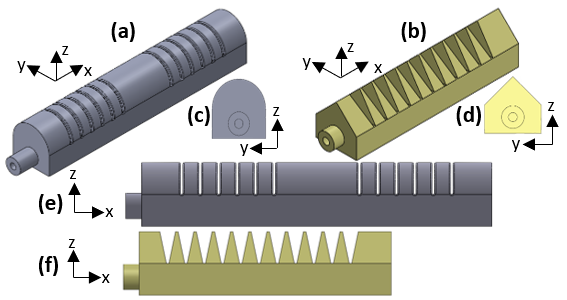}
\caption{Design changes made to the ventral fPN (yellow) from a traditional fPN design (gray). (a) Original fPN isometric view. (b) Ventral fPN isometric view. (c) Original fPN front view. (d) Ventral fPN front view. (e) Original fPN side view. (f) Ventral fPN side view.}
\label{fig.ventralfPN}
\end{figure}

The ventral fPN, shown in yellow in Fig.~\ref{fig.spasticAndProto}b, fits on the ventral aspect of the finger.
The design of the ventral fPN is detailed in Fig.~\ref{fig.ventralfPN}.
In comparison with a traditional fPN (Fig.~\ref{fig.ventralfPN} (gray)) meant for the dorsal aspect of the finger, the ventral fPN (Fig.~\ref{fig.ventralfPN} (yellow)) has four design changes to account for the reduced ventral space due to hyperflexed fingers in a spastic patient.
The first change is the reduction of the top half of the air chamber geometry in the z-x plane from rectangular to triangular, which is seen from the isometric views (Fig.~\ref{fig.ventralfPN}(a,b)) and the profile views (Fig.~\ref{fig.ventralfPN}(e,f)).
The second change is the removal of the solid regions between the air chambers, such that the entire length consists of triangular chambers.
These two changes allow the actuator to be passively compressed so that it can be squeezed into the reduced ventral space of a patient's hyperflexed finger.
The third change is a shorter length (Fig.~\ref{fig.ventralfPN}(e,f)), also accommodating for the reduced space in finger flexion.
The fourth change is the reduction of the top half of the air chamber geometry in the y-z plane from cylindrical to triangular (Fig.~\ref{fig.ventralfPN}(c,d)). 
This minimizes lateral space, allowing up to five actuators to fit on the ventral side of the hand.

Together, these changes are designed to allow the actuator to be compressed into the tight ventral space of a hyperflexed spastic finger.
The compressed state may be achieved through active robotic vacuum pressure or passive external forces from a human hand, either that of the patient's healthy hand or the hand of a caregiver. 
Once fitted, robotic positive pressure will expand the compressed actuator, encouraging finger extension and/or applying compression/massage therapy in tandem with the dorsal finger actuator.


\subsection{fPN fabrication}\label{subsec.fPNFabrication}

\begin{figure}[thpb]
\centering
\includegraphics[width=1\linewidth]{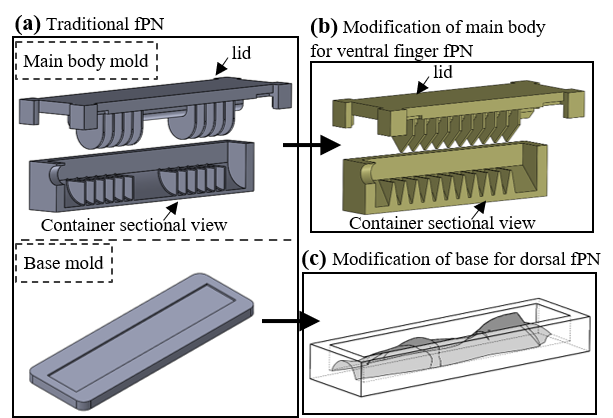}
\caption{Modifications of the dorsal and ventral fPNs from a traditional fPN. (a) Traditional fPN mold. (\emph{top}) Main body. (\emph{Bottom}) Base.
(b) Main body of ventral fPN is modified with triangular air chambers.
(c) Base of dorsal fPN is modified with finger scan topology.}
\label{fig.fPNMolds}
\end{figure}

The fPNs are fabricated through pour casting of silicone into 3D printed molds.  
The baseline mold assembly is shown in Fig.~\ref{fig.fPNMolds}a.
The molds are designed from a negative CAD model of the actuator.
All of the mold parts are 3D printed with polylactic acid (PLA).
The actuator is pour-casted in two parts: the main body and the base.  Once both parts are cured, they are sealed together with a third cast of a thin layer of silicone poured over the flat top of the base.

To fabricate the ventral fPN, the mold of the main body is modified to account for the triangular air chambers, shown in Fig.~\ref{fig.fPNMolds}b. 
Since the base of the ventral fPN is flat, the base mold remains the same, just modified in length.

\begin{figure}[thpb]
\centering
\includegraphics[width=1\linewidth]{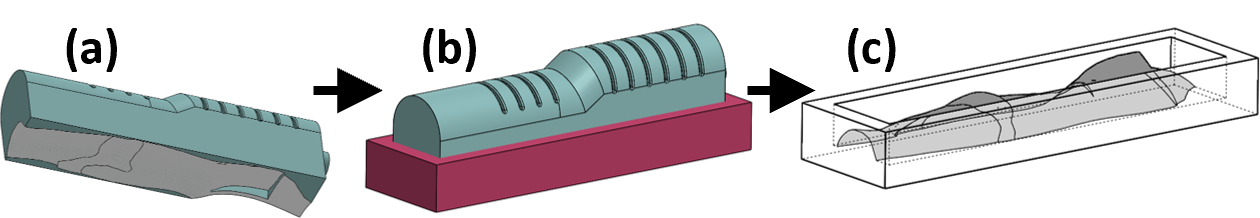}
\caption{Design of customized base mold for dorsal fPN. (a) Positive CAD model of dorsal fPN with finger topology. (b) Positive model is subtracted from a rectangular base. (c) Resulting base mold.
}
\label{fig.base}
\end{figure}

To fabricate the customized dorsal fPN, the main body remains the same, with bending air chamber regions centered at the MCP and PIP joints.
The base mold is modified as shown in Fig.~\ref{fig.fPNMolds}c to incorporate the topology of the dorsal aspect of the finger.
Fig.~\ref{fig.base}a is the positive CAD model of the actuator with the finger topology, presented in Section~\ref{subsec.customizedDorsalFpn}.
The positive model is then subtracted from a rectangular extrusion (Fig.~\ref{fig.base}b), which results in the final base mold (Fig.~\ref{fig.base}c).
The poured silicone takes the negative shape of the finger. 
The uncured silicone settles with gravity to a flat top surface, and once cured, a thin sealing layer is poured onto the flat surface to join it with the main body, as in a traditional fPN casting. 

\subsection{Palmar Bellow}\label{subsec.palmarBellow}
\begin{figure}[thpb]
\centering
\includegraphics[width=1\linewidth]{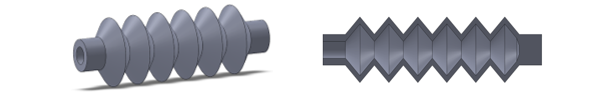}
\caption{(\emph{left}) Straight bellow model. (\emph{right}) Section view.}
\label{fig.straightBellowDesign}
\end{figure}

\setlength{\textfloatsep}{6pt}
The palmar bellow, shown in green in Fig.~\ref{fig.spasticAndProto}b, was iteratively designed starting from a traditional bellow-shaped pneumatic actuator, as shown in Fig.~\ref{fig.straightBellowDesign}.
The selection of a bellow-shaped pneumatic actuator for the palmar region was motivated by prior work demonstrating its suitability for wearable hand rehabilitation devices~\cite{hu2020novel}. Bellow geometries enable both extension and compression through controlled pressurization, a capability not achievable with many conventional soft actuators that typically provide unidirectional motion. This bidirectional behavior is particularly advantageous for applications involving spastic hands, where available workspace is limited, and the device must be compressed into a clenched, partially closed posture (see Fig.~\ref{fig.spasticAndProto}a). The ventral fPN presented in Section~\ref{subsec.ventralfPN} is designed to be compressed into the clenched finger, but can only be compressed on one side with bending. The bellow design here can achieve full linear compression and extension.

\begin{figure}[thpb]
\vspace{4pt}
\centering
\includegraphics[width=1\linewidth]{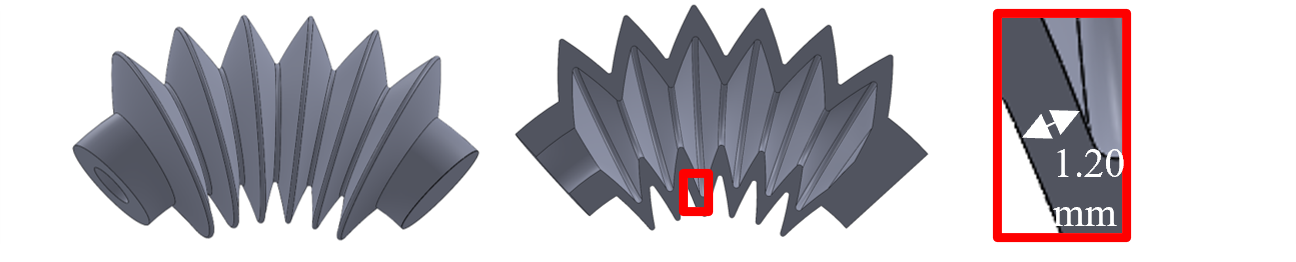}
\caption{(\emph{left}) Curved bellow model. (\emph{right}) Section view.}
\label{fig.curvedBellowDesign}
\end{figure}

\begin{figure}[thpb]
\centering
\includegraphics[width=1\linewidth]{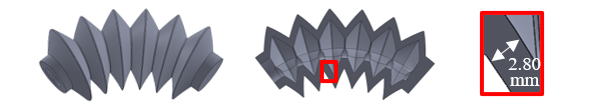}
\caption{(\emph{left}) Reinforced curved bellow. (\emph{right}) Section view.}
\label{fig.reinforcedBellowDesign}
\end{figure}

The straight bellow actuator (Fig.~\ref{fig.straightBellowDesign}) was modeled to establish baseline geometry that allows compression. A curved geometry (Fig.~\ref{fig.curvedBellowDesign}) was then generated to create a fan-like motion to move with the opposition and reposition of the palm. 
The bellow is designed to be compressed into the constrained space of the spastic clenched palm (see Fig.~\ref{fig.spasticAndProto}).
Curvature was applied using the SOLIDWORKS \textit{Flex} feature with a 90$^\circ$ bend, producing nonuniform wall thickness along the inner radius. Uniform wall thickness was restored by solidifying the lower portion of the bellow and applying a shell operation to match the upper wall thickness. Fillets were added at geometric transitions to reduce stress concentrations.
The reinforced curved bellow model with uniform wall thickness and fillets is shown in Fig.~\ref{fig.reinforcedBellowDesign}.

\begin{figure}[thpb]
\centering
\includegraphics[width=1\linewidth]{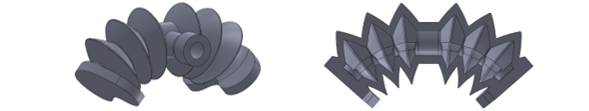}
\caption{(\emph{left}) Final palmar bellow actuator model, with flat sides and central air inlet. (\emph{right}) Section view.}
\label{fig.middleInletBellowDesign}
\end{figure}

The final bellow design shown in Fig.~\ref{fig.middleInletBellowDesign} is derived from the reinforced curved geometry. A centrally located air inlet was incorporated
to enable the terminal surfaces of the bellow to lie flat against the thenar and hypothenar eminences of the palm.
Once the bellow is compressed and fit into the clenched spastic palm, the terminal surfaces press against the palm.
These surfaces were widened and flattened to increase contact area with the palm and to apply more distributed, gentle pressure.
Upon pressurization, the actuator expands bidirectionally, generating outward forces that act on the palm to promote hand opening. 

\subsection{Wrist Base}\label{subsec.wristBase}
Using the scaled 3D hand model described in Section~\ref{subsec.customizedDorsalFpn}, a custom wrist base was designed and fabricated from thermoplastic polyurethane (TPU) as seen in blue in Fig.~\ref{fig.spasticAndProto}b. The wrist base consists of a separate top and bottom secured with a Velcro strap to facilitate donning and doffing.

A rigid attachment for the palmar bellow, 3D printed from PLA and shown in purple in Fig.~\ref{fig.spasticAndProto}b, is integrated into the wrist brace to hold the center of the bellow in a fixed ideal location.
The proximal end of the wrist brace incorporates channels to route and secure all pneumatic tubes coming from a remotely located pneumatic control station, improving organization and reducing interference during device operation.
	
\section{EXPERIMENTS \& RESULTS}\label{sec.experimentsAndResults}
We present the fabrication and finite element (FEM) analysis of the palmar bellow in Section~\ref{subsec.bellowIterations}, the FEM analysis of the dorsal and ventral fPN actuators in Section~\ref{subsec.fPNAnalysis}, and pneumatic experiments with the full glove assembled onto a participant's hand in Section~\ref{subsec.experimentalActuation}. 
While fPN fabrication is detailed in Section~\ref{sec.methods}, the palmar bellow fabrication is described here due to its additive manufacturing process, which differs from conventional molding and remains in the experimental phase.
The radial symmetry of the straight bellow or the taurus shape of the curved bellow, both with non-trivial internal geometries, would involve injection molding and lost wax casting if traditional techniques were used, which are time-consuming, laborious, and prone to failure.

\subsection{Bellow Iterations}\label{subsec.bellowIterations}

\begin{figure}[thpb]
\centering
\includegraphics[width=.9\linewidth]{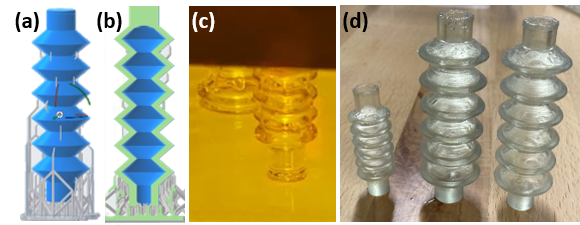}
\caption{SLA straight bellows: (a) Slicer output. (b) Slicer output section view showing no internal support. (c) Bellows printing on SLA printer. (d) Final printed straight bellows.}
\label{fig.slicerAndPrintingSLA}
\end{figure}

All bellow iterations (straight, curved, and center air inlet) were fabricated using a Formlabs Form 4L stereolithography printer with Elastic 50A V2 resin.  Prints were produced with a layer thickness of 0.1 mm using the default PreForm slicer settings. Support structures were initially generated automatically and restricted to external surfaces, after which additional supports were manually added to eliminate unsupported regions. The slicer output for these prints can be seen in Fig.~\ref{fig.slicerAndPrintingSLA}(a,b). Post-processing consisted of washing the printed parts in isopropyl alcohol using an agitator, followed by curing in a glass beaker with the parts fully submerged in room-temperature water. In accordance with the manufacturer’s guidelines, parts were initially cured in a Form Cure unit at 70 °C for 7 min; however, visual inspection indicated incomplete curing, as evidenced by residual green coloration. Consequently, the parts were returned to the Form Cure unit for an additional 10 min of curing at 70 °C, after which the support structures were removed.

Initial straight bellow iterations were experimentally evaluated to assess the effects of wall thickness and post-curing on pressure resistance.
A post-cured straight bellow with a wall thickness of 2.0 mm sustained pressurization for over 10 min without observable leakage or structural failure.
(See the supplementary video to view this experiment.)

\begin{figure}[thpb]
\vspace{2pt}
\centering
\includegraphics[width=.9\linewidth]{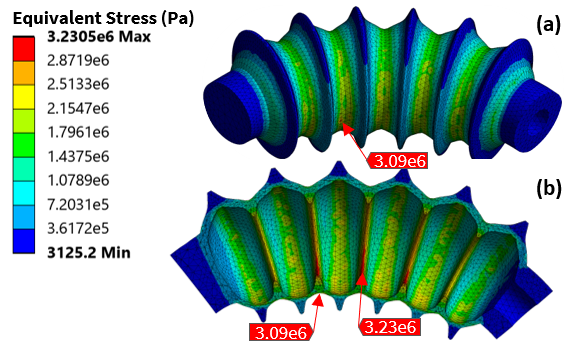}
\caption{Curved bellow FEM at 200 kPa with maximum stress locations. (a) Full model view. (b) Section view.}
\label{fig.curvedBellowFEM}
\end{figure}

The curved bellow iteration introduced additional mechanical challenges due to reduced wall thickness along the inner radius caused by the flex feature used during CAD modeling (see Section~\ref{subsec.palmarBellow}). A static FEM analysis was performed with Ansys Workbench to evaluate stress distribution and actuation behavior under a 200 kPa (29 psi) internal pressure load, with the inlet end fixed and gravity applied. The resulting equivalent stress and deformation profiles are shown in Fig.~\ref{fig.curvedBellowFEM}.

A compressible Neo-Hookean model was used to describe the material behavior of Elastic 50A Resin V2. The required hyperelastic parameters were estimated from manufacturer-reported data due to the lack of complete stress-strain measurements. The post-cured Shore A hardness of the material (55A) was converted to an effective small-strain Young’s modulus using an empirical exponential correlation between durometer hardness and elastic modulus for silicone-based elastomers~\cite{DowDurometer}:
\[E = C\,e^{kS}\]
where $S$ is the Shore A hardness. An upper-bound coefficient ($C = 0.10$~MPa, $k = 0.0235$) was selected to represent bulk tensile behavior, yielding $E \approx 3.6$~MPa. The material was assumed to be nearly incompressible, with a Poisson’s ratio of $\nu = 0.495$. The Neo-Hookean shear modulus was calculated as:
\[\mu = \frac{E}{2(1+\nu)} \approx 1.2~\text{MPa}\]
and the bulk modulus was calculated as:
\[K = \frac{E}{3(1-2\nu)} \approx 120~\text{MPa}\]
The compressibility parameter for the compressible Neo-Hookean formulation was then defined as $D_1 = 2/K$, giving $D_1 \approx 0.0167~\text{MPa}^{-1}$. These properties were used in finite element simulations and showed reasonable agreement with experimentally observed deformation and force--displacement trends for the tested geometries.

Simulation results predicted a maximum stress of 3.23~MPa, approaching the cured material strength of Elastic 50A V2 resin of 3.4~MPa. Peak stress was localized along the inner radius of the bellow, particularly on the outer surface farthest from the inlet, as shown in the section view of Fig.~\ref{fig.curvedBellowFEM}b, indicating a likely failure region.
In addition to these elevated stresses, the deformation profile exhibited elongation of the inner walls, rather than pure rotation (see supplementary video). 

Following simulation, the curved bellow was fabricated and then inflated with 50 mL of air to perform experimental testing. Structural failure occurred almost instantaneously at the inner radius region.
The observed failure closely matched the location of high stress predicted by the finite element analysis, validating the simulation results. This failure is attributed to stress concentration combined with reduced inner wall thickness resulting from the flex feature described in Section~\ref{subsec.palmarBellow}.

To address this limitation, the geometry was modified to reinforce the inner region of the bellow (see Section~\ref{subsec.palmarBellow}), which reduced the maximum predicted stress to 3.16~MPa (2.19\%) and shifted the peak stress location from the inner radius toward the mid-section of the bellow. Although the improvement to stress distribution was minimal, the reinforced geometry produced pure rotation behavior during actuation (see supplementary video). This controlled deformation profile improves suitability for palm-mounted compression and opening applications.

\begin{figure}[thpb]
\centering
\includegraphics[width=.9\linewidth]{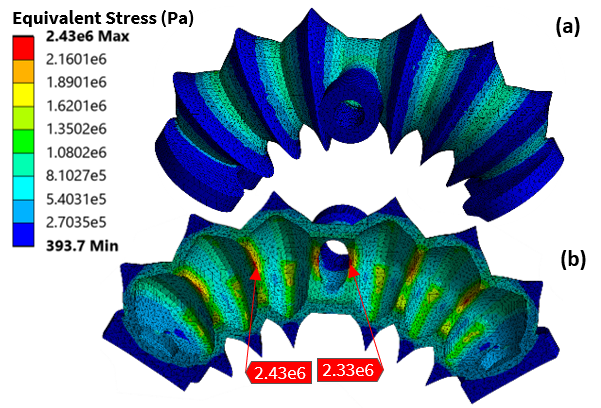}
\caption{Final palmar bellow FEM at 200 kPa. (a) Full view. (b) Section view with max stress locations.}
\label{fig.middleAirInletBellowFEM}
\end{figure}

Based on the findings from previous iterations, a middle air inlet bellow was designed to improve both mechanical performance and pneumatic integration. Finite element analysis was conducted under the same loading conditions used for prior designs, with a 200 kPa internal pressure applied to the bellow. The resulting equivalent stress and deformation profiles are shown in Fig.~\ref{fig.middleAirInletBellowFEM}. 

This geometry demonstrated substantial improvements in stress distribution and actuation behavior. The maximum stress experienced was 2.43~MPa, well below the cured material strength and representing a 26.1\% reduction compared to the previous curved bellow iteration. This significant stress reduction is attributed to the centrally located air inlet.
Since the central region had the peak stress, modifying this region to include the air inlet has the added benefit of decreasing the overall stress.

The deformation behavior closely matched the intended design objective, with the bellow remaining compact when unpressurized and fanning outward under applied pressure, as shown in Fig.~\ref{fig.middleAirInletBellowFEM}. This configuration allows the bellow to fit within the palm in its resting state and expand during actuation, enabling both palmar opening and localized compressive force delivery.

Following simulation, the middle air inlet bellow was fabricated and subjected to mechanical testing. The bellow was fully compressed using negative pressure and reinflated to its original state for five consecutive cycles. No leakage, structural damage, or material degradation was observed, indicating reliable performance under repeated loading.

\subsection{fPN Analysis}\label{subsec.fPNAnalysis}

\begin{figure}[thpb]
\centering
\includegraphics[width=1\linewidth]{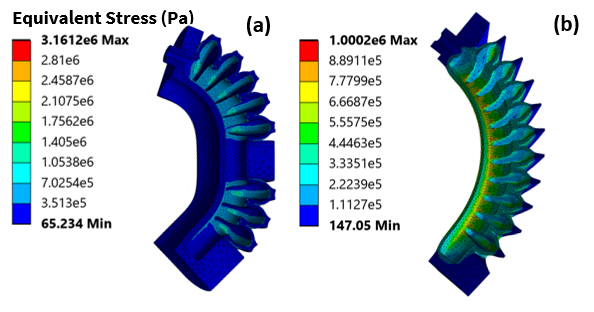}
\caption{Dorsal and ventral fPN FEM at 100 kPa. (a) Dorsal fPN section view. (b) Ventral fPN section view.}
\label{fig.femTest1}
\end{figure}

Similar FEM analyses were conducted for the fPN geometries with the results shown in Fig.~\ref{fig.femTest1}. The material behavior for this analysis was modeled as a hyperelastic solid using a third-order Yeoh model for Sorta Clear 40, with material parameters \( C_1 = 1.00 \times 10^{-1} \), \( C_2 = 2.11 \times 10^{-1} \), and \( C_3 = 1.66 \times 10^{-3} \), as reported in \cite{Marechal2021-SoRo}.
For the custom dorsal fPN, the geometry was simplified to run the FEM analysis. 
Instead of the fPN's contact surface being customized to the topology of the finger scan, the fPN was given a base conformal to a cylinder.
This reduced the number of facets of the geometry, allowing simulations to run smoothly.
Despite the simplification of the base geometry, there are still some aspects of customization in the longitudinal axis with regards to the placement of the air chambers and solid regions which align with the MCP and PIP joints and the bone regions, respectively.
In Fig.~\ref{fig.femTest1}a, the dorsal fPN stress analysis shows the maximum stress reaching $\sim$3.16~MPa, which is well below 5.51~MPa material stress limit for Sorta Clear 40 silicone.
The correct deformation behavior was observed,
demonstrating flexion as the internal air pressure increased.
The actuation of the dorsal fPN was stiffer compared to a traditional fPN with a flat base, both in FEM analysis and experimental testing.
This is due to the extra silicone in the customized base (see Fig.~\ref{fig.fPNMolds}), as well as in the simplified cylindrical form.
The geometry of the base is more difficult to compress compared to a flat base.
A future solution could be adding external cuts to the customized base centered at the joints, which will still allow for full conformal contact, while also creating more space for the base to compress and deform.

The ventral fPN analysis was performed with the exact geometry of the fabricated actuator.
Fig.~\ref{fig.femTest1}b shows the maximum stress of the actuator well below the 5.51 MPa stress limit, at only 1.00 MPa.
This reveals that the design changes made to the ventral fPN, as discussed in ~\ref{subsec.ventralfPN}, were successful. 
Not only does the fPN still actuate as expected, but it can do so without worries of material failure.

\subsection{Full glove Actuation}\label{subsec.experimentalActuation}

\begin{figure}[thpb]
\centering
\includegraphics[width=.9\linewidth]{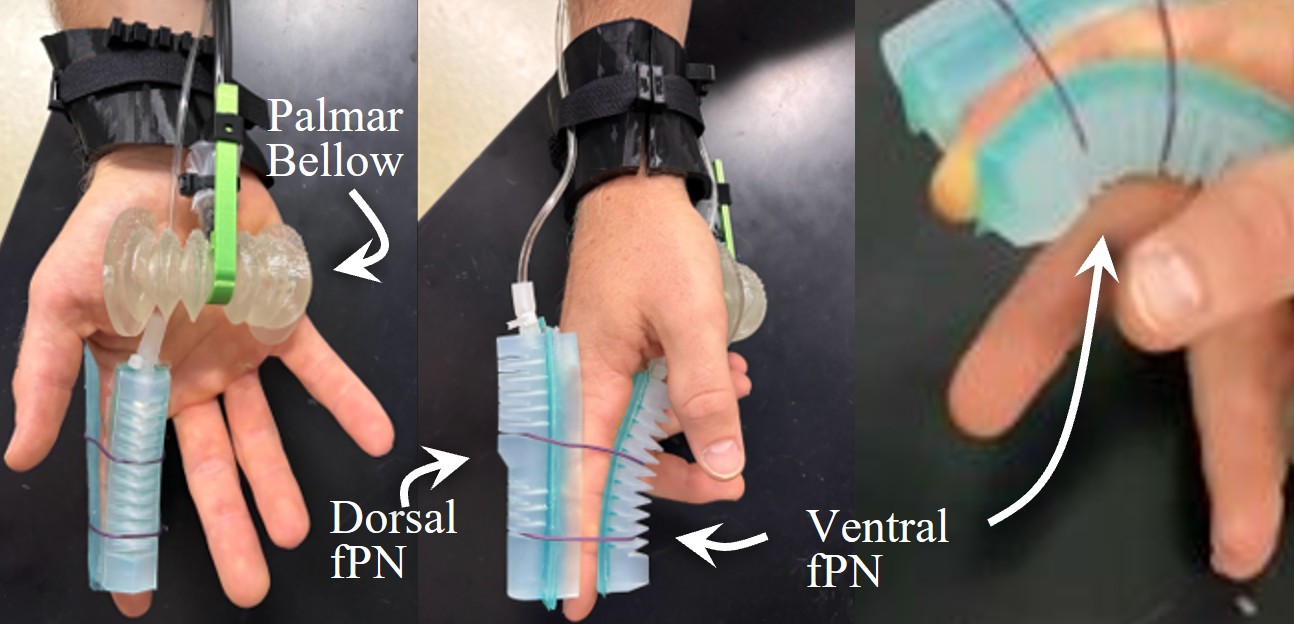}
\caption{Soft robotic exoglove experiments. \emph{(Closeup, right)} Ventral fPN is compressed into the flexed finger.}
\label{fig.fullAssemblyRealHand}
\end{figure}

The dorsal and ventral fPNs were evaluated alongside the palmar bellow on a healthy participant as shown in Fig.~\ref{fig.fullAssemblyRealHand}. For the index finger experiment with the fPNs, the participant brought their finger into a flexed position, mimicking finger hyperflexion often observed in patients suffering from spasticity. The ventral fPN was compressed manually and placed inside the reduced ventral space of the finger. The dorsal fPN was then placed on the dorsal side of the finger with temporary straps to connect the fPNs together.
The trends observed in FEM held true in experimental testing. The ventral fPN was easily compressed to fit in the tight space on the ventral side of the flexed finger. The stiffer dorsal fPN was difficult to deform onto the dorsal side of the flexed finger. 
The actuators were robotically inflated with a pneumatic control station~\cite{Massoud2025-RAL,Massoud2024-IEEEAccess} running PID pressure control, applying a reference ramp from 0 to 20 kPa, allowing for a compressive massage with gradually increasing force. 
When the ventral fPN was pressurized, the user felt a force encouraging the finger to extend, even against some resistance. 
The user reported that simultaneous pressurization of the ventral and dorsal fPNs felt like a massage-type compression that relaxed the muscles.

Following evaluation of the fPN actuators, the palmar bellow was assessed on the same participant. The participant positioned the thumb in opposition toward the fifth digit to approximate palmar contraction observed in individuals with spasticity. The palmar bellow was manually compressed and placed within the palmar region of the hand, after which positive pneumatic pressure was applied. Inflation of the bellow generated outward forces on thenar and hypothenar eminences of the palm, producing a clearly perceptible opening force. The experiments of the ventral fPN and palmar bellow can be viewed in the supplementary video.

\section{CONCLUSION}\label{sec.conclusion}
This work presents an early-stage prototype of a pneumatically actuated soft robotic exoglove intended to support pain reduction and mitigate hand spasticity. The device integrates three actuators: a dorsal fPN, a ventral fPN, and a palmar bellow actuator.
The device operates through the coordinated action of a dorsal and ventral fPN, which apply opposing pressures to the finger to produce a compression-like massage, and a palmar bellow actuator that assists with hand opening while providing distributed pressure to the palm. The forces applied by the actuators could be used to provide resistance for hand strengthening and assistance for neural plasticity rehabilitation, in addition to reducing spasticity.
The dorsal fPN is customized to conform to the topological anatomy of the hand, and the ventral fPN is fabricated to compress inside hyperflexed fingers. The palmar bellow actuator, manufactured using stereolithography (SLA) printing, is designed to be compressed into the palmar region of a contracted hand.
	
Several limitations were identified.
The topological customization of the dorsal fPN added thickness to its base, creating unwanted stiffness in the joint regions.  External cuts in the base could be introduced to mitigate this effect.
The palmar bellow was difficult to compress due to the stiffness of the SLA elastic resin, whereas thinner walls resulted in silicone resin delamination and tearing.
This indicates fundamental restrictions in SLA 3D printing for highly deformable soft pneumatic actuators.
Future designs will explore softer silicone resins and other 3D printing technologies.
A strapping mechanism to join the dorsal and ventral fPNs will also be designed.  Current results are based on anecdotal feedback from a limited sample size; future work will involve exploratory clinical studies with a larger and more diverse population. 

\bibliographystyle{unsrt}
\bibliography{references}

\end{document}